\documentclass[letterpaper,10pt,conference]{ieeeconf}

\IEEEoverridecommandlockouts
\usepackage{graphicx}
\usepackage{booktabs}
\usepackage[table]{xcolor}
\definecolor{avgcolumn}{RGB}{236,242,248}
\definecolor{tablegroup}{RGB}{242,244,247}
\definecolor{avgintersection}{RGB}{218,230,240}
\usepackage{multirow}
\usepackage{amsmath}
\usepackage{amssymb}
\usepackage{cite}
\usepackage{url}
\usepackage{microtype}
\usepackage{placeins}
\usepackage{balance}

\newcommand{\realpravg}{86.7}
\newcommand{\realprgain}{8.7}

\makeatletter
\let\paper@default@makecaption\@makecaption
\renewcommand{\fnum@table}{Table~\thetable}
\long\def\@makecaption#1#2{%
  \ifx\@captype\@IEEEtablestring
    \par\noindent
    \parbox{\linewidth}{%
      \normalfont\small\raggedright\textbf{#1.}\enspace #2\par}%
    \vskip 4pt
  \else
    \paper@default@makecaption{#1}{#2}%
  \fi}
\makeatother

\title{\LARGE \bf
PACT-WAM: Predicting Actions and Visual Foresight with Compact Temporal Encoding for Robot Manipulation
}
\author{Yushan Liu$^{1,*}$, Jingjing Fan$^{1,*}$, Shoujie Li$^{2}$, Yifan Xie$^{1}$, Xiao-Ping Zhang$^{1}$, \textit{Fellow, IEEE}, Wenbo Ding$^{1,3,\dagger}$%
\thanks{$^{*}$These authors contributed equally to this work.}%
\thanks{$^{\dagger}$Corresponding author.}%
\thanks{$^{1}$Tsinghua University}%
\thanks{$^{2}$Nanyang Technological University}%
\thanks{$^{3}$Xspark AI}
}

\begin{document}
\bstctlcite{PACT:reference-control}

\maketitle
\thispagestyle{empty}
\pagestyle{empty}
\raggedbottom

\begin{abstract}
Robot manipulation uses temporal context to select actions and visual foresight to assess their consequences, yet dense representations of past and future observations incur substantial processing costs. We introduce PACT-WAM, a world--action model that jointly generates a 16-step action trajectory and its temporally corresponding visual forecast through conditional flow sampling. Hierarchical history encoding assigns coarse spatial representations to earlier observations and finer representations to recent ones, retaining 16 observations with 256 tokens per view---75\% fewer than dense encoding of the same frames. A shared flow module jointly updates continuous action and visual states through two modality-specific heads under transition-wise causal attention, and a TiTok-VAE decoder reconstructs multi-view future images from the visual latents. Decoded forecasts also support Proposal Review (PR), a vision--language model component for execution-prefix selection and proposal rejection. Without PR, PACT-WAM achieves average success rates of 98.6\%, 92.3\%, and 78.0\% on LIBERO, RoboTwin 2.0, and real-world Piper tasks, respectively. PR provides a test-time enhancement, raising these rates to 99.5\%, 93.4\%, and \realpravg\%. Ablations show that hierarchical history allocation and joint action--visual generation improve control success, while analyses of visual capacity and forecast-guided execution characterize the trade-offs between success and proposal-generation cost.
\end{abstract}

% !TeX root = root.tex
\section{Introduction}

Large-scale robot data and vision--language pretraining enable generalist manipulation policies to follow natural-language instructions~\cite{black2410pi0,intelligence2025pi05}. Manipulation also uses two forms of temporal information: history provides context about motion and prior interactions~\cite{fvla_memoryvla,fvla_contextvla,fvla_sdvla}, while visual foresight predicts state changes for assessing proposed actions~\cite{fvla_videovla,fvla_dreamzero}. World--action models connect history, actions, and future observations, but both past and future images are costly. Encoding 16 history frames at 64 tokens each requires 1,024 tokens per view; multi-step forecasts add a separate visual-latent sequence. Reducing their repeated scene content must preserve the temporal information needed for action selection and execution review.

\begin{figure}[t]
  \vspace*{6pt}
\centering
  \includegraphics[width=\columnwidth]{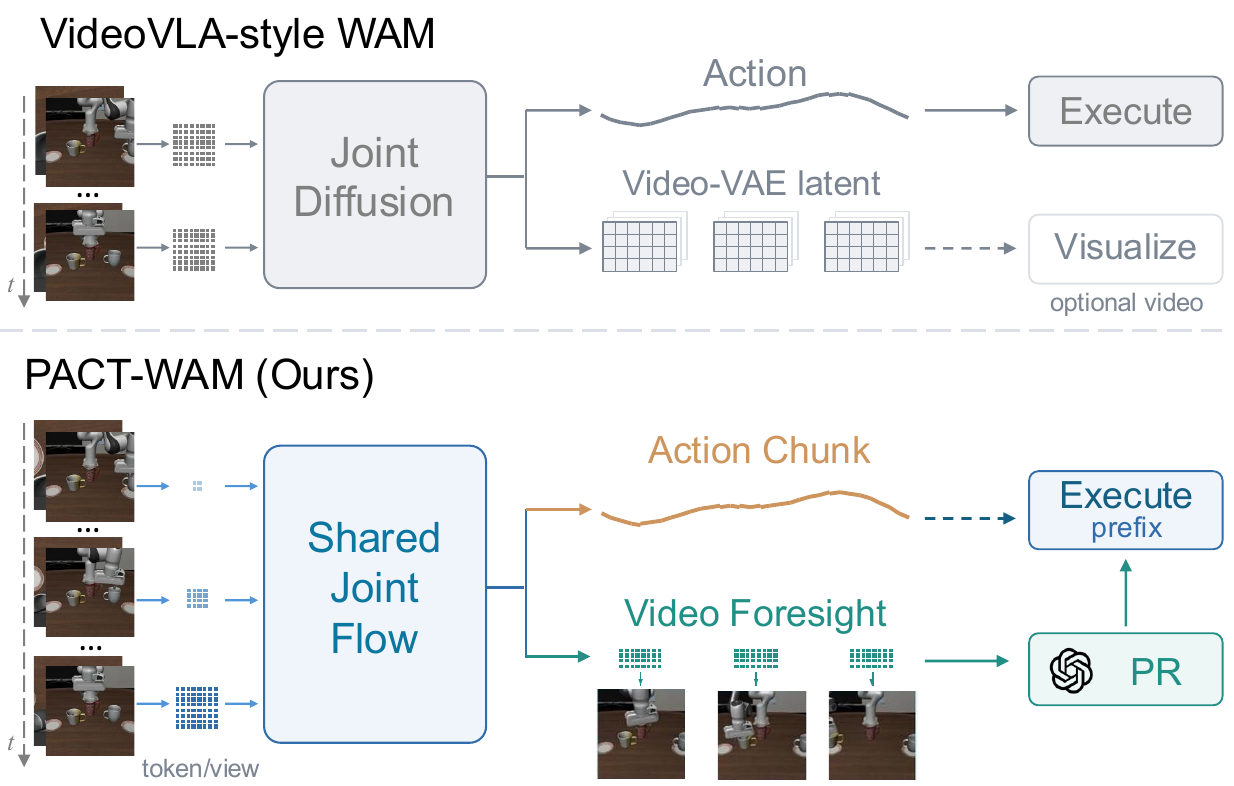}

  \caption{\textbf{Paradigm comparison.} Top: dense-history joint action--video generation incurs high representation costs, while visual forecasts remain separate from execution decisions. Bottom: PACT-WAM combines hierarchical history encoding with compact visual latents to jointly generate temporally paired actions and forecasts, using Proposal Review (PR) to guide execution.}
  \label{fig:overview}

\end{figure}

Existing methods address different parts of this challenge. History compression, token pruning, and active perception improve visual inputs for action prediction~\cite{fvla_contextvla,fvla_sdvla,fvla_vlapruner,liu2025avr}. Modular visual foresight couples a predictor with a separate controller~\cite{fvla_unipi,fvla_susie,chen2025robohorizon,liu2026checkvla}, while unified world--action models couple action generation and visual prediction~\cite{zhu2025unified,bi2025motus,li2026lingbotva}. Video codecs with temporal downsampling group multiple physical steps within each latent time slice. Other efficient designs omit future generation or decoding at inference, limiting visual evidence for review~\cite{chen2026lawam,yuan2026fastwam}. The problem is: how can we design compact representations that preserve historical coverage and an independently decodable state at each proposed transition, together with a protocol that uses these forecasts to guide execution?

We propose \textbf{PACT-WAM} (Fig.~\ref{fig:overview}), which jointly generates 16 actions and corresponding multi-view forecasts from compact history. Three choices connect representation efficiency to execution review. First, hierarchical encoding assigns coarse representations to older frames and finer representations to recent frames, retaining 16 observations with 256 tokens per view. Second, each future image uses 32 continuous TiTok-VAE latents without temporal downsampling, pairing every action with an independently decodable subsequent visual state. Third, a shared flow transformer with transition-wise causal attention and two velocity heads jointly updates action and visual states, exchanging information during sampling. Proposal Review (PR) uses four parallel vision--language model requests to assess nested forecast prefixes. The controller combines consecutive approvals with cross-review vetoes to select an execution horizon, or jointly resamples an unapproved proposal within a bounded retry budget.

Our contributions are as follows:
\begin{itemize}
    \item We introduce recency-based history allocation with 256 tokens per view. On LIBERO, it outperforms uniform encoding of the same 16 frames at equal token count (98.6\% versus 87.5\%) and achieves comparable success to dense encoding (98.0\%) with 75\% fewer history tokens.
    \item We propose \textbf{PACT-WAM}, which jointly generates actions and visual states in compact per-image latent space, providing an independently decodable forecast at each physical transition. PR uses these forecasts for execution-prefix approval, cross-review vetoes, and bounded joint resampling.
    \item PACT-WAM achieves 98.6\%, 92.3\%, and 78.0\% success without PR on LIBERO, RoboTwin 2.0, and Piper, respectively; test-time PR raises these to 99.5\%, 93.4\%, and \realpravg\%. Ablations evaluate history allocation, joint generation, visual capacity, and the overall PR protocol, including proposal-generation costs.
\end{itemize}

\section{Related Work}

\subsection{VLA Policies and Temporal Context}

Vision--language--action (VLA) policies map observations and instructions to robot actions. Alongside visual--haptic demonstration collection, pretrained action modeling learns control through continuous action generation, policy fine-tuning, or auxiliary reasoning and dynamics supervision~\cite{chao2025exoviha,black2410pi0,intelligence2025pi05,kim2025fine,li2026zr0,li2026dypesvla}. History aggregation supplies temporal context through recurrent states, memory banks, or compressed multi-frame representations~\cite{fvla_roboflamingo,fvla_memoryvla,fvla_contextvla,fvla_sdvla}. Efficient inference reduces computation through visual token pruning and caching, compact action tokenization, or streaming action decoding~\cite{fvla_vlapruner,xu2025vla,pertsch2025fast,li2026flashvla}. Under a limited visual budget, longer temporal coverage must be balanced against spatial detail. PACT-WAM allocates coarse representations to older observations and finer representations to recent ones within a fixed history-token budget.

\subsection{Modular Visual Foresight for Control}

Modular visual foresight connects a visual predictor to a separate policy or controller. Visual-goal planning translates predicted images or videos into actions through inverse dynamics, goal-conditioned policies, or geometric control~\cite{fvla_unipi,fvla_susie,fvla_avdc}. Action-conditioned world models support simulation, policy refinement, proposal evaluation, or execution monitoring~\cite{chen2025robohorizon,jiang2025world4rl,qi2025strengthening,liu2026checkvla}. Both routes require coordinating prediction with a separate control component to realize visual goals or assess candidate actions. PACT-WAM couples action trajectories and visual futures in a joint conditional flow, allowing their representations to exchange information during sampling from shared history.

\subsection{Unified and Efficient World--Action Models}

Unified world--action models learn action generation and visual prediction within a shared framework. Multimodal generative modeling couples both tasks through causal, diffusion, or flow architectures~\cite{cen2025rynnvla,zhu2025unified,fvla_videovla,fvla_dreamzero,bi2025motus,li2026lingbotva,liu2026oawam}. Efficient foresight uses compact future features, image-level prediction, or omits future generation or decoding during policy inference~\cite{chen2026lawam,yang2026lilawam,zhang2026imagewam,yuan2026fastwam,liu2026perceptdrive}. These designs expose a trade-off between inference cost and the availability of visual forecasts for execution decisions. PACT-WAM combines hierarchical history with compact, decodable per-image latents, jointly sampling temporally paired actions and forecasts that Proposal Review uses to select an execution prefix or reject the proposal.

\begin{figure*}[t]
  \vspace*{6pt}
\centering

    \includegraphics[width=\textwidth]{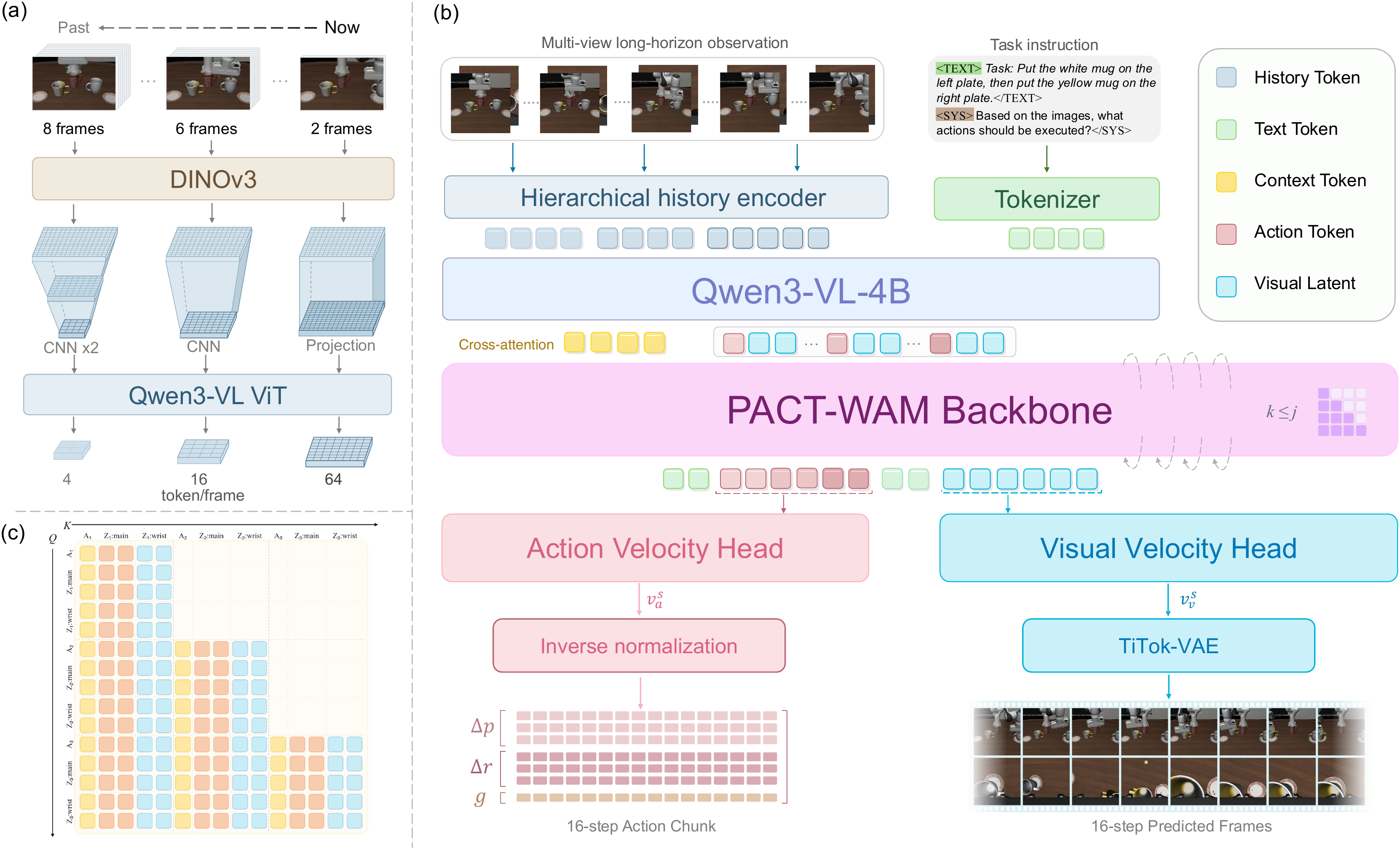}

    \caption{\textbf{PACT-WAM architecture.} \textbf{(a) Hierarchical history encoder.} DINOv3 features undergo recency-dependent compression and Qwen visual processing, yielding 4/16/64 tokens per image for the oldest eight, middle six, and latest two observations. \textbf{(b) Joint action--visual generation.} Sixteen multi-view history observations and a language instruction form fixed Qwen context. Given this context, noisy action/visual states, and flow time, the shared backbone predicts velocities through separate heads. Joint integration, inverse normalization, and TiTok-VAE decoding yield 16 actions and multi-view forecasts of the state after each action. \textbf{(c) Transition-wise causal attention mask.} In this subset, rows/columns denote queries/keys; colored cells allow the same or earlier transitions, and blank cells block later ones.}
    \label{fig:future_vla}

\end{figure*}

\section{Method}
\label{sec:method}

We propose PACT-WAM, a world--action model for robot manipulation. PACT-WAM combines hierarchical history encoding and joint action--visual flow generation with Proposal Review for execution-prefix selection and joint resampling (Fig.~\ref{fig:future_vla}).

\subsection{Problem Formulation}
\label{sec:formulation}

Let $O_u=(I_u^1,\ldots,I_u^V)$ denote RGB observations from $V$ cameras at physical time step $u$. At decision time $t$, the input is an instruction $\ell$ and ordered history $\mathcal H_t=(O_{\tau_1},\ldots,O_{\tau_N})$, with $\tau_1\leq\cdots\leq\tau_N=t$. We jointly model
\begin{equation}
\begin{aligned}
 A_t&=(a_t,\ldots,a_{t+T-1}),\\
 O_t^+&=(O_{t+1},\ldots,O_{t+T}).
\end{aligned}
\label{eq:io}
\end{equation}
Here $a_u\in\mathbb R^{d_a}$ is a platform-specific action. The $j$th pair, $(a_{t+j-1},O_{t+j})$, associates an action with the observation after the first $j$ actions. The forecast count is $F=T=16$, with a default history length of $N=16$. PR uses this action--observation correspondence to review candidate execution prefixes (Sec.~\ref{sec:inference}).

The history encoder supplies final visual tokens $C_t$ and intermediate DeepStack features to Qwen3-VL-4B, which combines them with instruction $\ell$ to obtain shared context $h_t$. Each future image is represented by a continuous TiTok-VAE latent $Z_j^v=E_v(I_{t+j}^v)\in\mathbb R^{K\times d_v}$, with $K=32$ and $d_v=16$. The latent encoder uses the posterior mean to define a deterministic training target. We stack these latents into $Z_t\in\mathbb R^{T\times V\times K\times d_v}$. With training-set normalization maps $\mathcal N_a$ and $\mathcal N_v$, the joint generation state is
\begin{equation}
 X_t=(\bar A_t,\bar Z_t)
 =\bigl(\mathcal N_a(A_t),\mathcal N_v(Z_t)\bigr).
\label{eq:joint_state}
\end{equation}
Conditioned on $h_t$, the joint flow sampler updates action and visual states together across all 16 paired steps, with transition-wise causal attention encoding their dependence across physical time.

\subsection{Hierarchical History Encoding}
\label{sec:arch}

The history pathway detailed in Fig.~\ref{fig:future_vla}(a) uses per-image spatial compression selected by observation recency. Frozen DINOv3 ViT-B/16~\cite{simeoni2025dinov3} supplies a $16\times16\times768$ patch grid, excluding the classification and four register tokens. Three independently parameterized CNN branches $Q_r$ produce $r\times r\times1024$ features for $r\in\{4,8,16\}$: two stride-2 convolutions, one stride-2 convolution, and a pointwise projection, respectively. These branches provide both spatial compression and channel projection into the Qwen vision width; DINOv3 and the CNNs replace Qwen's pixel patch embedding while retaining its visual Transformer.

The retained visual pathway adds interpolated spatial position embeddings and uses two-dimensional rotary positions. Its learned merger reduces the number of positions by four and projects features into the language hidden dimension~\cite{bai2025qwen3vltechnicalreport}. Denoting this visual pathway and final merger by $\Phi$, with $D_i^v=E_{\mathrm{DINO}}(I_{\tau_i}^v)$,
\begin{equation}
\begin{aligned}
 C_i^v&=\Phi\bigl(Q_{r_i}(D_i^v)\bigr),\\
 C_t&=\operatorname{Concat}_{i,v}C_i^v.
\end{aligned}
\label{eq:history_encoding}
\end{equation}
Qwen's DeepStack mergers additionally project intermediate visual features into early language layers when computing $h_t$. History images are serialized in time order with a fixed camera order at each time; image boundaries and Qwen multimodal rotary positions organize the context. Cross-frame interactions occur in the language backbone after per-image encoding. For $N=16$, the oldest eight, middle six, and latest two observations use $r=4,8,16$, yielding 4, 16, and 64 tokens per image after merging. The history budget per view is
\begin{equation}
 B_h=8\cdot4+6\cdot16+2\cdot64=256.
\label{eq:history_budget}
\end{equation}
This allocation preserves all 16 observations with 75\% fewer history tokens than encoding each at 64 tokens.

\begin{figure*}[t]
  \vspace*{6pt}
\centering

 \includegraphics[width=\textwidth]{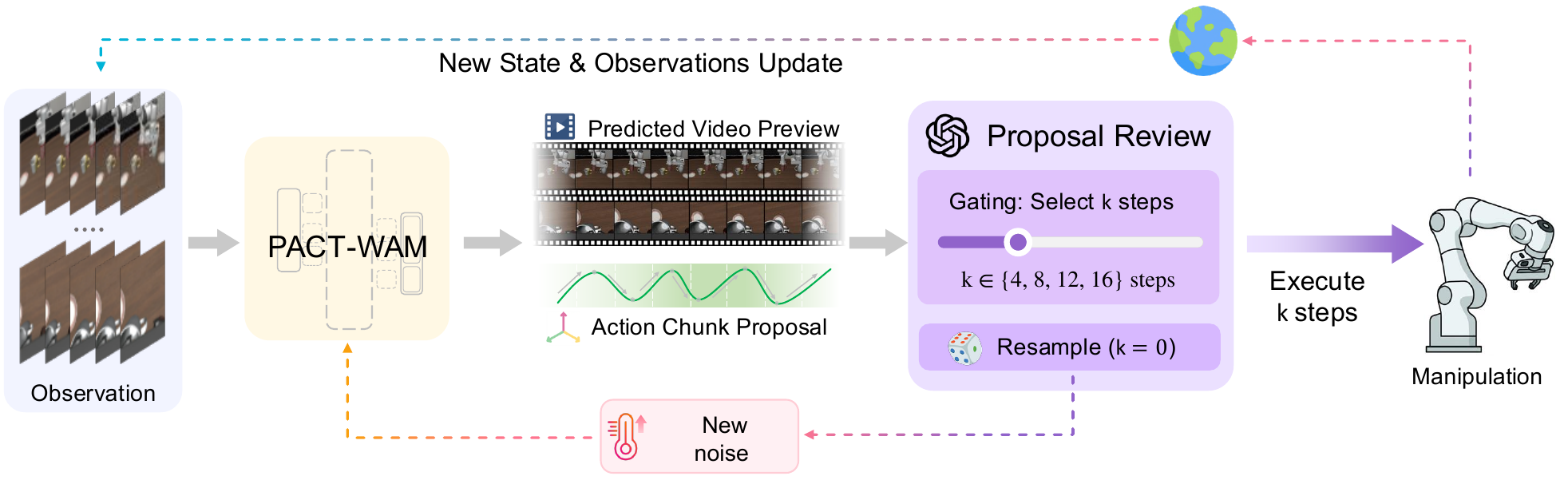}

 \caption{\textbf{Proposal Review (PR) for PACT-WAM.} Four parallel VLM requests review prefixes at 4, 8, 12, and 16 steps. The controller selects the largest consecutively approved prefix permitted by all reported vetoes. Rejected proposals are jointly resampled, with at most three retries; incomplete review withholds execution.}
 \label{fig:proposal_review}

\end{figure*}

\subsection{Temporally Aligned Joint Flow Generation}
\label{sec:tokenization}

\paragraph*{Continuous representation and coupled heads}
We use a pretrained continuous TiTok-VAE encoder and its matching decoder~\cite{yu2024image,kim2025democratizing}, with $K=32$ latent positions per image by default. The codec encodes each image independently and shares weights across views, preserving a visual representation at every forecast step. There are $L_z=TVK$ visual latent positions, each with $d_v=16$ continuous channels. Actions are normalized continuous vectors, with one action position per physical step. Both modalities carry the same transition index $j$ for the pair in \eqref{eq:io}; modality, view, and latent-slot embeddings distinguish their remaining positions.

The shared flow transformer $F_\theta$, shown as the PACT-WAM backbone in Fig.~\ref{fig:future_vla}(b), receives the noisy action and visual states, a flow-time embedding $s\in[0,1]$, and the fixed context $h_t$. We group its output positions by physical transition: block $j$ contains action $a_{t+j-1}$ and every visual latent position for $O_{t+j}$ across all views. Self-attention uses a block-causal visibility mask,
\begin{equation}
 M_{jk}=\begin{cases}
 1,&k\le j,\\
 0,&k>j.
 \end{cases}
\label{eq:temporal_mask}
\end{equation}
In every flow-transformer layer, the mask lets a query in block $j$ attend to all positions in block $k$ with $M_{jk}=1$, coupling action and visual positions within each transition and propagating information from earlier transitions. The mask spans all $T=16$ transitions and all latent slots. Fig.~\ref{fig:future_vla}(c) illustrates Q--K visibility for three transitions, with one action position and two of the $K=32$ latent slots per view. Both modalities separately cross-attend to all valid history--language context positions, held fixed throughout sampling. Separate heads $g_a$ and $g_v$ read the corresponding action and visual hidden positions:
\begin{equation}
\begin{aligned}
 (H_a^s,H_v^s)&=F_\theta(\bar A_t^s,\bar Z_t^s,s;h_t,M),\\
 (v_a^s,v_v^s)&=\bigl(g_a(H_a^s),g_v(H_v^s)\bigr).
\end{aligned}
\label{eq:joint_heads}
\end{equation}
The two heads predict velocities in normalized action and visual-latent spaces, coupling the modalities through their shared dependence on noisy states within the permitted temporal prefix. Flow time $s$ parameterizes generation progress, while $j$ indexes physical transitions along the proposed trajectory.

\paragraph*{Training}
We use the linear-path velocity regression objective of conditional flow matching~\cite{lipman2023flowmatching}, with the velocity network constrained by \eqref{eq:temporal_mask}. For each paired training target $X_t$, we draw independent standard Gaussian action and visual noises, concatenate them as $\epsilon$, and use one common $s\sim\mathcal U[0,1]$:
\begin{equation}
 X_t^s=(1-s)\epsilon+sX_t,\qquad u_t=X_t-\epsilon.
\label{eq:flow_path}
\end{equation}
Let $n_a=Td_a$ and $n_v=TVKd_v$ count the scalar entries of each target. The loss averages within each modality before combining the two terms:
\begin{equation}
 \mathcal L_{\mathrm{FM}}
 =\mathbb E_{X_t,\epsilon,s}\left[
 \sum_{m\in\{a,v\}}\frac{\lambda_m}{n_m}
 \left\|v_m^s-u_{t,m}\right\|_F^2\right].
\label{eq:flow_loss}
\end{equation}
The mask imposes a temporal dependency prior on the velocity field; this restricted regression does not guarantee physical causality or exact transport along the unrestricted CFM path. We optimize this objective conditioned on history and language, with $\lambda_a=\lambda_v=1$ by default. The pretrained DINOv3 and TiTok-VAE encoder/decoder remain frozen, while the compression branches, Qwen context pathway, flow transformer, and two output heads are trained with the world--action objective.

\paragraph*{Joint sampling and decoding}
At inference, a fresh joint noise state is integrated with the learned velocity field:
\begin{equation}
 \frac{dX_t^s}{ds}=v_\theta(X_t^s,s;h_t,M),
 \qquad X_t^0=\epsilon.
\label{eq:flow_sampling}
\end{equation}
Here $v_\theta=(v_a^s,v_v^s)$ combines the shared transformer and the two heads in \eqref{eq:joint_heads}. At each evaluation on the fixed integration schedule, the masked transformer predicts velocities for all 16 transitions, and the solver jointly advances the action and visual states over the complete horizon. From the terminal state $X_t^1=(\hat{\bar A}_t,\hat{\bar Z}_t)$, we recover the final outputs as
\begin{equation}
\begin{aligned}
 \hat A_t&=\mathcal N_a^{-1}(\hat{\bar A}_t),\\
 \hat I_{t+j}^v&=D_v\!\left(
 [\mathcal N_v^{-1}(\hat{\bar Z}_t)]_j^v\right).
\end{aligned}
\label{eq:joint_decoding}
\end{equation}
Inverse normalization restores physical action units and the visual-latent scale, after which the TiTok-VAE decoder $D_v$ reconstructs the multi-view images. Both inverse normalization and the TiTok-VAE decoder precede image decoding.

\begin{table*}[t]
  \vspace*{6pt}
\centering
\caption[Baseline comparison on LIBERO and RoboTwin 2.0]{\textbf{Baseline success rates (\%) on LIBERO and RoboTwin 2.0.} PACT-WAM with and without PR is compared against VLA and WAM baselines. \textbf{Bold} and \underline{underlined} values mark the highest and second-highest success rates per column, including ties.}
\label{tab:main_results}
\begingroup
\fontsize{9.5}{11.4}\selectfont
\renewcommand{\arraystretch}{1.15}
\setlength{\tabcolsep}{4pt}
\setlength{\aboverulesep}{0pt}
\setlength{\belowrulesep}{0pt}
\newcommand{\sr}[1]{\makebox[2.6em][r]{#1}}
\newcommand{\liberoavg}[1]{\sr{#1}\kern\cmidrulekern}
\newcommand{\mainmetricwidth}{\dimexpr(\textwidth-.355\textwidth-18\tabcolsep)/7\relax}
\begin{tabular}{w{l}{.25\textwidth}*{4}{w{c}{\mainmetricwidth}}>{\columncolor{avgcolumn}}w{c}{\mainmetricwidth}>{\columncolor{white}[\dimexpr\tabcolsep+\cmidrulekern\relax][\tabcolsep]}w{c}{\mainmetricwidth}w{c}{.105\textwidth}>{\columncolor{avgcolumn}}w{c}{\mainmetricwidth}}
\toprule[0.9pt]
\multirow{2}{*}{\textbf{Method}} & \multicolumn{5}{c}{\textbf{LIBERO}} & \multicolumn{3}{c}{\textbf{RoboTwin 2.0}} \\
\cmidrule(lr){2-6}\cmidrule(l){7-9}
& Spatial & Object & Goal & Long & \textbf{Avg.}\kern\cmidrulekern & Clean & Randomized & \textbf{Avg.} \\
\midrule
\rowcolor{tablegroup}\multicolumn{5}{l}{\textit{Vision-Language-Action Methods}} & \cellcolor{avgintersection} & \multicolumn{2}{>{\columncolor{tablegroup}[\dimexpr\tabcolsep+\cmidrulekern\relax][\tabcolsep]}c}{} & \cellcolor{avgintersection} \\
$\pi_0$~\cite{black2410pi0} & \sr{96.8} & \sr{98.8} & \sr{95.8} & \sr{85.2} & \liberoavg{94.2} & \sr{65.9} & \sr{58.4} & \sr{62.2} \\
$\pi_{0.5}$~\cite{intelligence2025pi05} & \sr{\underline{98.8}} & \sr{98.2} & \sr{98.0} & \sr{92.4} & \liberoavg{96.9} & \sr{82.7} & \sr{76.8} & \sr{79.8} \\
ZR-0~\cite{li2026zr0} & \sr{97.4} & \sr{99.4} & \sr{98.0} & \sr{96.4} & \liberoavg{97.8} & \sr{88.7} & \sr{88.0} & \sr{88.3} \\
DyPES-VLA~\cite{li2026dypesvla} & \sr{\underline{98.8}} & \sr{99.4} & \sr{97.0} & \sr{96.8} & \liberoavg{98.0} & \sr{88.8} & \sr{89.3} & \sr{89.0} \\
FlashVLA ($\pi_{0.5}$, sync.)~\cite{li2026flashvla} & \sr{98.6} & \sr{99.0} & \sr{97.8} & \sr{96.2} & \liberoavg{97.9} & \sr{90.8} & \sr{90.2} & \sr{90.5} \\
\rowcolor{tablegroup}\multicolumn{5}{l}{\textit{World-Action Model Methods}} & \cellcolor{avgintersection} & \multicolumn{2}{>{\columncolor{tablegroup}[\dimexpr\tabcolsep+\cmidrulekern\relax][\tabcolsep]}c}{} & \cellcolor{avgintersection} \\
Motus~\cite{bi2025motus} & \sr{96.8} & \sr{\underline{99.8}} & \sr{96.6} & \sr{97.6} & \liberoavg{97.7} & \sr{88.7} & \sr{87.0} & \sr{87.8} \\
LiLa-WAM~\cite{yang2026lilawam} & \sr{98.0} & \sr{98.8} & \sr{97.2} & \sr{94.2} & \liberoavg{97.1} & \sr{90.5} & \sr{89.0} & \sr{89.8} \\
LaWAM~\cite{chen2026lawam} & \sr{\textbf{99.4}} & \sr{99.6} & \sr{98.4} & \sr{97.0} & \liberoavg{\underline{98.6}} & \sr{92.6} & \sr{89.8} & \sr{91.2} \\
Fast-WAM~\cite{yuan2026fastwam} & \sr{98.2} & \sr{\textbf{100.0}} & \sr{97.0} & \sr{95.2} & \liberoavg{97.6} & \sr{91.9} & \sr{91.8} & \sr{91.8} \\
LingBot-VA~\cite{li2026lingbotva} & \sr{98.5} & \sr{99.6} & \sr{97.2} & \sr{\underline{98.5}} & \liberoavg{98.5} & \sr{\underline{92.9}} & \sr{91.6} & \sr{92.2} \\
\midrule
\rowcolor{tablegroup}\textbf{PACT-WAM (w/o PR)} & \sr{98.4} & \sr{99.6} & \sr{\underline{99.4}} & \sr{97.0} & \cellcolor{avgintersection}\liberoavg{\underline{98.6}} & \sr{92.7} & \sr{\underline{91.9}} & \cellcolor{avgintersection}\sr{\underline{92.3}} \\
\rowcolor{tablegroup}\textbf{PACT-WAM (w/ PR)} & \sr{\textbf{99.4}} & \sr{\textbf{100.0}} & \sr{\textbf{100.0}} & \sr{\textbf{98.6}} & \cellcolor{avgintersection}\liberoavg{\textbf{99.5}} & \sr{\textbf{94.0}} & \sr{\textbf{92.8}} & \cellcolor{avgintersection}\sr{\textbf{93.4}} \\
\bottomrule[0.9pt]
\end{tabular}
\endgroup
\end{table*}

\begin{figure*}[t]
  \vspace*{6pt}
\centering
    \includegraphics[width=\textwidth]{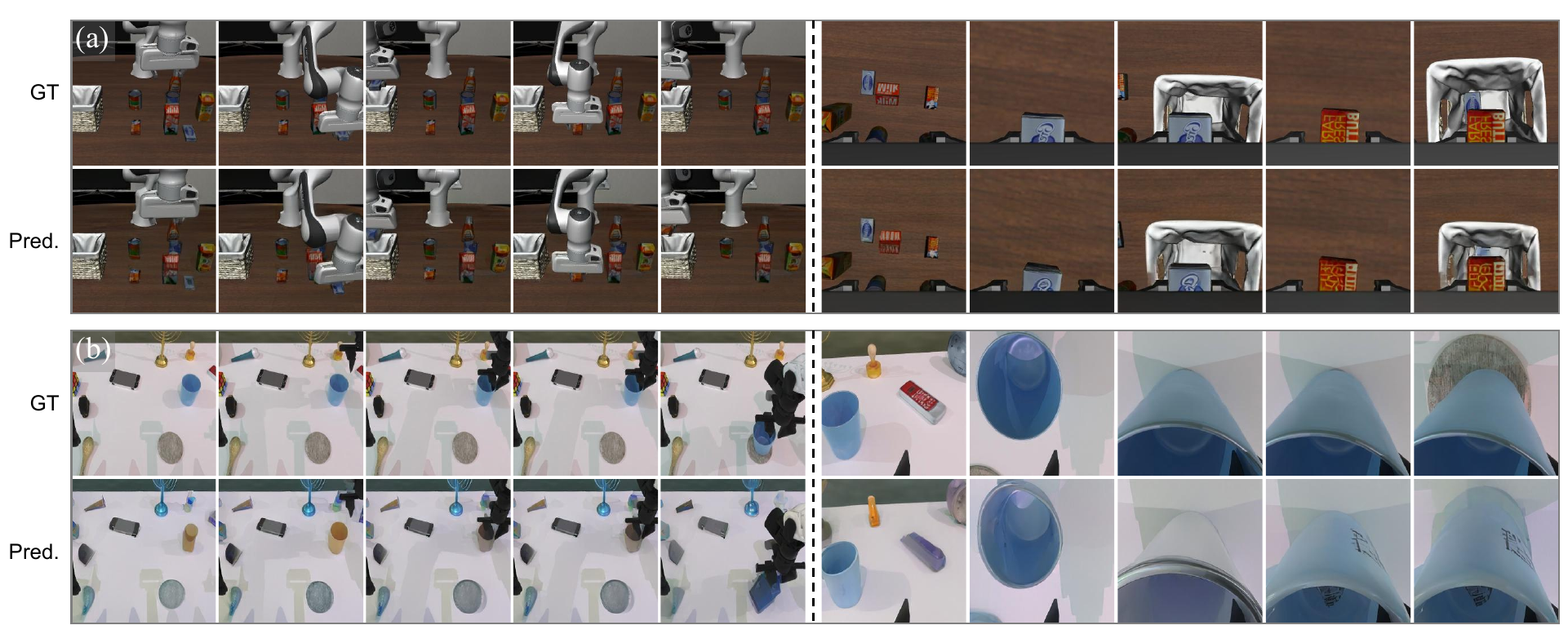}

    \caption{\textbf{PACT-WAM rollouts in simulation.} (a) LIBERO; (b) RoboTwin 2.0. The dashed divider separates external (left) and wrist (right) views. Each view pairs observed keyframes (GT, upper row) with predicted future frames (Pred., lower row), ordered from left to right.}
    \label{fig:sim_rollout}

\end{figure*}

\subsection{Proposal Review}
\label{sec:inference}

PR reviews each joint proposal in one batch of four parallel GPT-5.5 requests (Fig.~\ref{fig:proposal_review}). Request $h\in\mathcal P=\{4,8,12,16\}$ receives the shared instruction, current observations, and previews at checkpoints $4,8,\ldots,h$, with all views at each checkpoint. It assesses task-consistent progress and visible failures over that prefix, including object misalignment, unstable grasping, collisions, and implausible transitions; task completion by $h$ is not required. Each returns a status $r_{t,h}\in\{\mathrm A,\mathrm R,\mathrm U\}$ (approve, reject, unknown). Rejection or uncertainty also specifies the earliest blocked checkpoint $b_{t,h}\in\mathcal P\cap[4,h]$, defaulting to 4 if its range is unclear; approval sets $b_{t,h}=\infty$. The controller selects the largest consecutively approved prefix below every reported veto:
\begin{equation}
 \begin{aligned}
 \beta_t&=\min_{h\in\mathcal P}b_{t,h},\\
 k_t&=4\sum_{i=1}^{4}\prod_{j=1}^{i}
 \mathbf 1[r_{t,4j}=\mathrm A,\;4j<\beta_t].
 \end{aligned}
\label{eq:proposal_review}
\end{equation}
Positive $k_t$ requires four valid responses by the deadline. A valid veto at checkpoint 4 permits early $k_t=0$; subsequent cancellations are not timeouts. The robot waits for review, executes the approved prefix, and replans. Valid $k_t=0$ triggers up to three joint resamples under unchanged context and solver settings; four unapproved proposals terminate the episode. Technical exceptions withhold execution; episodes terminated by exceptions count as failures.

% Queue Fig. 5 before the page-5 experiment introduction for page 6.
\begin{figure*}[t]
  \vspace*{6pt}
\centering
    \includegraphics[width=\textwidth]{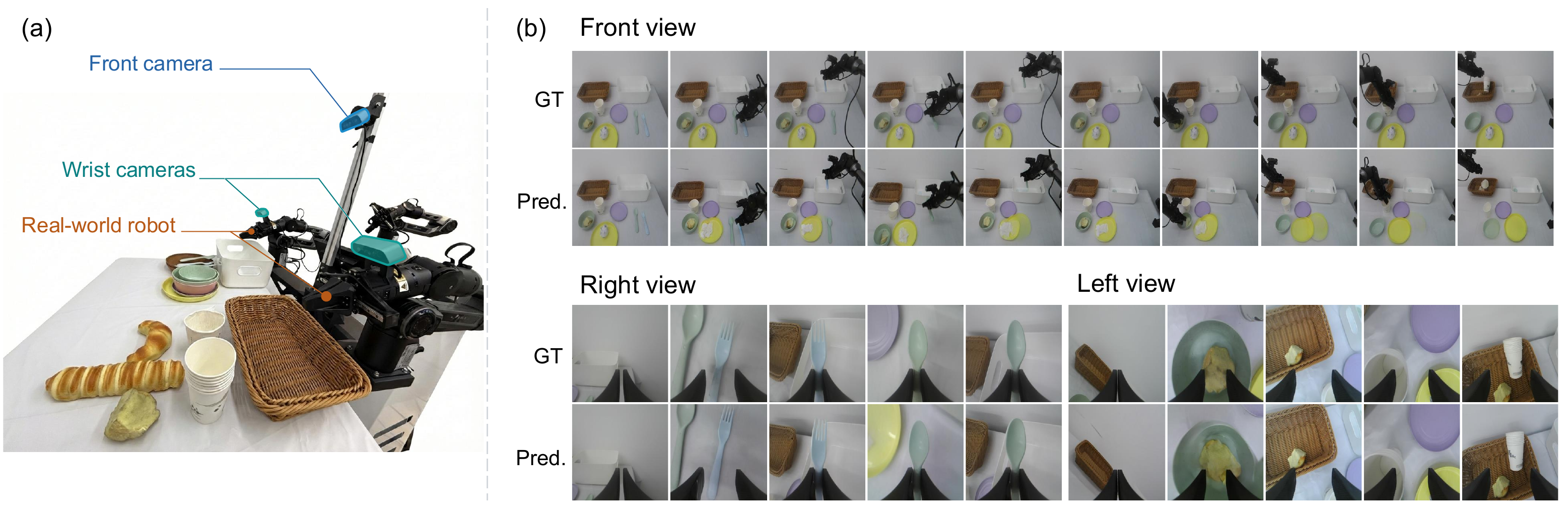}

    \caption{\textbf{Real-world setup and PACT-WAM rollout.} (a) AgileX Piper dual-arm platform with a front (overhead) camera and two wrist cameras. (b) Front (top), right wrist (bottom left), and left wrist (bottom right) views. Each view pairs observed (GT, upper row) and predicted (Pred., lower row) keyframes, ordered from left to right.}
    \label{fig:real_rollout}

\end{figure*}

\section{Experiments}
\label{sec:experiments}
\setcounter{dbltopnumber}{2}

We evaluate PACT-WAM in simulation and on a real robot, then analyze history allocation, joint generation, visual capacity, and PR.

\subsection{Benchmarks and Experimental Setup}
\label{sec:benchmark}

\paragraph*{Simulation Benchmarks}

We train one PACT-WAM policy per simulation benchmark. For LIBERO~\cite{liu2023libero}, we use all 1,693 training trajectories in the processed OpenPi LeRobot dataset to jointly train on 40 tasks across Spatial, Object, Goal, and Long, evaluating 50 episodes per task and aggregating successes over 500 episodes per suite. For RoboTwin 2.0~\cite{chen2025robotwin}, we jointly train on all 50 bimanual tasks using 27,500 trajectories: 2,500 Clean and 25,000 Randomized (50 and 500 per task, respectively). The resulting policy is evaluated with unseen instructions over 100 episodes per task in each condition. Randomized evaluation varies scene appearance and object configurations. Dataset counts refer to complete demonstration trajectories.

\paragraph*{Real-world Experiments}
We evaluate three tasks on the AgileX Piper platform (Fig.~\ref{fig:real_rollout}(a)). \textit{Target-Oriented Sorting} places designated objects from clutter into assigned receptacles; \textit{Collaborative Handover} transfers an object between arms through a shared region before placement; \textit{Table Cleanup} combines both skills to clear multiple objects.

We use 600 demonstrations: 150 each for Sorting and Handover, and 300 for Cleanup, with a 9:1 trajectory-level train/validation split per task. Each method jointly adapts one policy to all three tasks using identical splits; PACT-WAM variants share a checkpoint. We evaluate 50 independent episodes per task and method with matched initial configurations and instructions. Success requires complete task fulfillment, judged from observed outcomes independently of PR; Avg. is the unweighted mean across tasks. Physical human assistance or exhaustion of the PR retry budget counts as failure.

\begin{table}[t]
  \vspace*{6pt}
\centering
  \caption{\textbf{Real-world success rates (\%) on Piper.} \textbf{Bold} and \underline{underlined} values denote the best and second-best results in each column, including ties.}
  \label{real-table}
  \begingroup
  \fontsize{9}{10.8}\selectfont
  \renewcommand{\arraystretch}{1.15}
  \setlength{\tabcolsep}{3pt}
  \setlength{\aboverulesep}{0pt}
  \setlength{\belowrulesep}{0pt}
  \newcommand{\sr}[1]{\makebox[2.6em][r]{#1}}
  \begin{tabular}{w{l}{.373\columnwidth}w{c}{.118\columnwidth}w{c}{.154\columnwidth}w{c}{.132\columnwidth}>{\columncolor{avgcolumn}}w{c}{\dimexpr.223\columnwidth-10\tabcolsep\relax}}
    \toprule[0.9pt]
    \multirow{2}{*}{\textbf{Method}} & \multicolumn{4}{c}{\textbf{Real-world (Piper)}} \\
    \cmidrule(l){2-5}
    & Sorting & Handover & Cleanup & \textbf{Avg.} \\
    \midrule
    \rowcolor{tablegroup}\multicolumn{4}{l}{\textit{VLA Methods}} & \cellcolor{avgintersection} \\
    OpenVLA-OFT~\cite{kim2025fine} & \sr{68.0} & \sr{26.0} & \sr{6.0} & \sr{33.3} \\
    $\pi_0+$FAST~\cite{pertsch2025fast} & \sr{74.0} & \sr{40.0} & \sr{22.0} & \sr{45.3} \\
    $\pi_{0.5}$~\cite{intelligence2025pi05} & \sr{92.0} & \sr{\underline{72.0}} & \sr{58.0} & \sr{74.0} \\
    \rowcolor{tablegroup}\multicolumn{4}{l}{\textit{WAM Methods}} & \cellcolor{avgintersection} \\
    Fast-WAM~\cite{yuan2026fastwam} & \sr{88.0} & \sr{64.0} & \sr{48.0} & \sr{66.7} \\
    LaWAM~\cite{chen2026lawam} & \sr{94.0} & \sr{68.0} & \sr{54.0} & \sr{72.0} \\
    \midrule
    \rowcolor{tablegroup}\textbf{PACT-WAM (w/o PR)} & \sr{\underline{98.0}} & \sr{\underline{72.0}} & \sr{\underline{64.0}} & \cellcolor{avgintersection}\sr{\underline{78.0}} \\
    \rowcolor{tablegroup}\textbf{PACT-WAM (w/ PR)} & \sr{\textbf{100.0}} & \sr{\textbf{82.0}} & \sr{\textbf{78.0}} & \cellcolor{avgintersection}\sr{\textbf{\realpravg}} \\
    \bottomrule[0.9pt]
  \end{tabular}

  \endgroup
\end{table}

% Display the visual diagnostics across both columns for arXiv.
\begin{figure*}[t]
  \vspace*{6pt}
\centering
  \includegraphics[width=\textwidth]{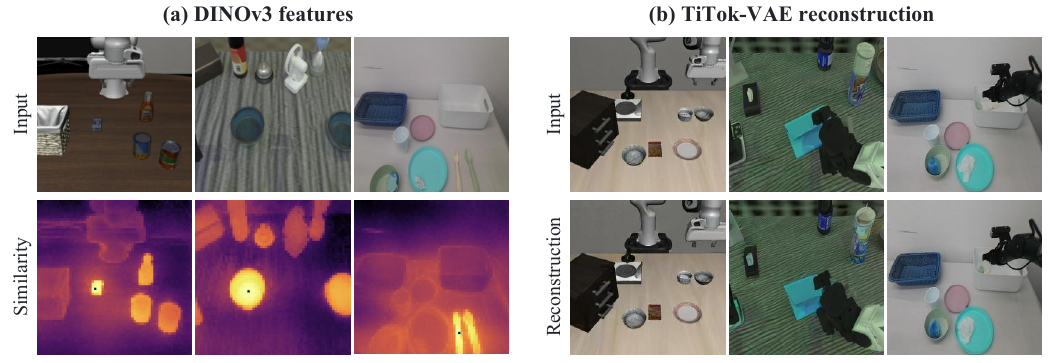}

  \caption{\textbf{Spatial features and visual reconstruction.} (a) Input observations (top) and DINOv3 feature-similarity maps (bottom). Warmer colors indicate higher similarity to the marked query patch. (b) Input observations (top) and TiTok-VAE reconstructions (bottom), using 32 continuous latent positions per image.}
  \label{fig:representation_diagnostics}

\end{figure*}

\subsection{Implementation Details}
\label{sec:implementation}

\paragraph*{Model Configuration}
The history frontend and visual codec use $256\times256$ RGB inputs, $N=T=F=16$, and 256 history tokens per view (Sec.~\ref{sec:arch}). The flow transformer has 16 layers, hidden width 1,024, and 16 attention heads. We use 10 Euler steps per proposal (10 velocity-network evaluations).

\paragraph*{Training Configuration}
The trainable modules in Sec.~\ref{sec:tokenization} undergo full-parameter optimization using AdamW with $\beta=(0.9,0.999)$, $\epsilon=10^{-8}$, and weight decay 0.1. We train for two epochs with an effective global batch size of 256. The learning rate warms up linearly for 1,000 optimizer updates to $10^{-4}$, then follows cosine decay.

\paragraph*{Data Construction}
Data are split at the trajectory level before constructing training windows, which advance by one recorded frame. Short histories repeat the earliest available observation. Each target requires 16 valid actions and 16 future observations; incomplete terminal targets are excluded. Action and visual-latent normalization statistics are fitted on the training split and held fixed for evaluation.

\paragraph*{Evaluation Protocol}
We evaluate the same joint sampler with full 16-action chunks (w/o PR) or with PR (Sec.~\ref{sec:inference}). LIBERO uses its default action interface and control frequency. Proposal latency and peak memory are measured on an NVIDIA L20 using bf16. Single-sample proposal latency runs from observation receipt to physical-unit actions and, for joint variants, decoded RGB previews; it includes history encoding, context computation, flow sampling, inverse normalization, and any RGB decoding. Review, rejected proposals, resampling, and execution waits are excluded; the reported latency measures proposal generation. History and generation ablations omit PR (Sec.~\ref{sec:ablation}).

External baseline scores in Table~\ref{tab:main_results} retain their published protocols and provide benchmark context: the $\pi_0$/$\pi_{0.5}$ scores follow ImageWAM~\cite{zhang2026imagewam}, and Motus's LIBERO scores follow LaWAM~\cite{chen2026lawam}. FlashVLA uses synchronous $\pi_{0.5}$ execution ($d=0$); LingBot-VA's Easy/Hard settings map to Clean/Randomized. Success rates are rounded to one decimal; LIBERO Avg. is the unweighted mean over all four suites; RoboTwin Avg. averages Clean and Randomized before rounding.

\begin{table*}[t]
  \vspace*{6pt}
\centering
\caption{\textbf{History allocation and compression on LIBERO.} (a) Varying history allocation at a fixed budget of 256 tokens per view; (b) varying compression with 16 history frames. Allocations count frames encoded at 4/16/64 tokens per image, from oldest to newest.}
\label{tab:history_encoding}
\begingroup
\fontsize{9}{10.8}\selectfont
\renewcommand{\arraystretch}{1.12}
\setlength{\tabcolsep}{3pt}
\setlength{\aboverulesep}{0pt}\setlength{\belowrulesep}{0pt}
\begin{minipage}[t]{.445\textwidth}
\centering
\textbf{(a) Fixed budget: 256 tokens per view}\par\vspace{4pt}
\begin{tabular}{w{l}{\dimexpr.36\linewidth-2\tabcolsep\relax}w{c}{\dimexpr.12\linewidth-2\tabcolsep\relax}w{c}{\dimexpr.25\linewidth-2\tabcolsep\relax}w{c}{\dimexpr.27\linewidth-2\tabcolsep\relax}}
\toprule
Allocation & $N$ & Success (\%) $\uparrow$ & Mem. (GB) \\
\midrule
0:0:4 (dense) & 4 & 90.2 & 23.2 \\
0:8:2 & 10 & \underline{95.8} & 23.3 \\
0:16:0 (uniform) & 16 & 87.5 & 23.4 \\
\rowcolor{tablegroup}8:6:2 (default) & 16 & \textbf{98.6} & 23.4 \\
16:4:2 & 22 & 87.6 & 23.4 \\
32:0:2 & 34 & 79.1 & 23.3 \\
\bottomrule
\end{tabular}
\end{minipage}\hfill
\begin{minipage}[t]{.535\textwidth}
\centering
\textbf{(b) Fixed history: $N=16$ frames}\par\vspace{4pt}
\begin{tabular}{w{l}{\dimexpr.28\linewidth-2\tabcolsep\relax}w{c}{\dimexpr.12\linewidth-2\tabcolsep\relax}w{c}{\dimexpr.21\linewidth-2\tabcolsep\relax}w{c}{\dimexpr.22\linewidth-2\tabcolsep\relax}w{c}{\dimexpr.17\linewidth-2\tabcolsep\relax}}
\toprule
Allocation & Tokens & Success (\%) $\uparrow$ & Mem. (GB) & Time (s)\\
\midrule
16:0:0 & 64 & 40.1 & 20.7 & 0.9 \\
12:4:0 & 112 & 73.8 & 21.9 & 0.8 \\
0:16:0 (uniform) & 256 & 87.5 & 23.4 & 1.1 \\
\rowcolor{tablegroup}8:6:2 (default) & 256 & \textbf{98.6} & 23.4 & 1.1 \\
8:4:4 & 352 & 97.6 & 25.7 & 1.4 \\
0:0:16 (dense) & 1,024 & \underline{98.0} & 38.3 & 1.4 \\
\bottomrule
\end{tabular}
\end{minipage}
\endgroup

\end{table*}

\begin{table*}[t]
  \vspace*{6pt}
\centering
\begin{minipage}[c]{\dimexpr.5\textwidth-.5\columnsep\relax}
\centering
\caption{\textbf{Joint generation and visual capacity on LIBERO.} (a) Action-only, visual-auxiliary, and joint variants without PR; (b) forecast quality.}
\label{tab:joint_foresight}
\begingroup
\fontsize{8.5}{10.2}\selectfont
\renewcommand{\arraystretch}{1.08}
\setlength{\tabcolsep}{3pt}
\setlength{\aboverulesep}{0pt}\setlength{\belowrulesep}{0pt}
\newcommand{\metricheader}[2]{\makebox[\dimexpr#1\columnwidth-2\tabcolsep\relax][c]{#2}}
\textbf{(a) Control and per-proposal cost}\par\vspace{2pt}
\begin{tabular}{w{l}{\dimexpr.44\columnwidth-2\tabcolsep\relax}w{r}{\dimexpr.22\columnwidth-2\tabcolsep\relax}w{r}{\dimexpr.16\columnwidth-2\tabcolsep\relax}w{r}{\dimexpr.18\columnwidth-2\tabcolsep\relax}}
\toprule
Variant & \metricheader{.22}{Success (\%)} & \metricheader{.16}{Time (s)} & \metricheader{.18}{Mem. (GB)} \\
\midrule
Action only & 93.6 & 0.7 & 11.4 \\
Visual auxiliary, $K=32$ & 96.4 & 0.8 & 11.6 \\
\rowcolor{tablegroup}Joint, $K=32$ (default) & \textbf{98.6} & 1.1 & 23.4 \\
Joint, $K=64$ & \underline{97.3} & 1.4 & 27.8 \\
\bottomrule
\end{tabular}
\par\vspace{4pt}\textbf{(b) Forecast quality}\par\vspace{2pt}
\begin{tabular}{w{l}{\dimexpr.38\columnwidth-2\tabcolsep\relax}w{r}{\dimexpr.15\columnwidth-2\tabcolsep\relax}w{r}{\dimexpr.15\columnwidth-2\tabcolsep\relax}w{r}{\dimexpr.16\columnwidth-2\tabcolsep\relax}w{r}{\dimexpr.16\columnwidth-2\tabcolsep\relax}}
\toprule
Method & \metricheader{.15}{PSNR $\uparrow$} & \metricheader{.15}{SSIM $\uparrow$} & \metricheader{.16}{LPIPS $\downarrow$} & \metricheader{.16}{FVD $\downarrow$} \\
\midrule
RynnVLA-002~\cite{cen2025rynnvla} & \underline{21.88} & 74.44 & 21.54 & 503.74 \\
\rowcolor{tablegroup}PACT-WAM, $K=32$ & 21.82 & \underline{79.92} & \underline{11.65} & \underline{32.85} \\
PACT-WAM, $K=64$ & \textbf{22.12} & \textbf{80.34} & \textbf{11.31} & \textbf{31.90} \\
\bottomrule
\end{tabular}
\endgroup
\end{minipage}\hfill
\begin{minipage}[c]{\dimexpr.5\textwidth-.5\columnsep\relax}
\centering
\caption{\textbf{Proposal review on RoboTwin 2.0 Randomized.} (a) Reviewer-input diagnostics on paired proposals; (b) closed-loop success and mean approved horizon under fixed-horizon execution and PR.}
\label{tab:review_analysis}
\begingroup
\fontsize{8.5}{10.2}\selectfont
\renewcommand{\arraystretch}{1.08}
\setlength{\tabcolsep}{2pt}
\setlength{\aboverulesep}{0pt}\setlength{\belowrulesep}{0pt}
\newcommand{\metricheader}[2]{\makebox[\dimexpr#1\columnwidth-2\tabcolsep\relax][c]{#2}}
\textbf{(a) Reviewer input}\par\vspace{2pt}
\begin{tabular}{w{l}{\dimexpr.70\columnwidth-2\tabcolsep\relax}w{r}{\dimexpr.15\columnwidth-2\tabcolsep\relax}w{r}{\dimexpr.15\columnwidth-2\tabcolsep\relax}}
\toprule
Reviewer input & \metricheader{.15}{FA (\%) $\downarrow$} & \metricheader{.15}{FR (\%) $\downarrow$} \\
\midrule
Current observation only & \underline{24.0} & 16.0 \\
Copied current frames & 31.0 & \underline{15.0} \\
\rowcolor{tablegroup}Predicted previews & \textbf{12.0} & \textbf{9.0} \\
\bottomrule
\end{tabular}
\par\vspace{4pt}\textbf{(b) Closed-loop execution}\par\vspace{2pt}
\begin{tabular}{w{l}{\dimexpr.44\columnwidth-2\tabcolsep\relax}w{r}{\dimexpr.28\columnwidth-2\tabcolsep\relax}w{r}{\dimexpr.28\columnwidth-2\tabcolsep\relax}}
\toprule
Protocol & \metricheader{.28}{Success (\%) $\uparrow$} & \metricheader{.28}{\shortstack{Mean approved\\horizon}} \\
\midrule
Full chunk (w/o PR) & 91.9 & 16.0 \\
Fixed-12 & 92.0 & 12.0 \\
Fixed-6 & \underline{92.4} & 6.0 \\
\rowcolor{tablegroup}PR (ours) & \textbf{92.8} & 14.0 \\
\bottomrule
\end{tabular}
\endgroup
\end{minipage}

\end{table*}

\subsection{Task Results}
\label{sec:evaluation_results}

\paragraph*{LIBERO}
Without PR, PACT-WAM achieves 98.6\% average success on LIBERO, matching LaWAM's reported average (Table~\ref{tab:main_results}). Test-time PR raises success to \textbf{99.5\%}, including 100.0\% on Object and Goal.

\paragraph*{RoboTwin 2.0}
The base policy achieves 92.3\% average success across 50 bimanual tasks (Table~\ref{tab:main_results}). PR raises this to \textbf{93.4\%} (+1.1 percentage points), with 94.0\% on Clean and 92.8\% on Randomized.

\paragraph*{Real-world manipulation}
On Piper, the base policy achieves 78.0\% average success, exceeding $\pi_{0.5}$'s 74.0\% by 4.0 percentage points (Table~\ref{real-table}). PR raises success to \textbf{\realpravg\%} (117/150 to 130/150; +\realprgain\ points), with 100.0\%, 82.0\%, and 78.0\% on Sorting, Handover, and Cleanup.

\subsection{Ablation and Diagnostic Analysis}
\label{sec:ablation}

History and generation ablations use LIBERO without PR; PR analyses use RoboTwin Randomized. Comparisons within each ablation share training data, evaluation seeds, and sampling settings. Shading marks defaults; Mem. and Time denote peak inference memory and per-proposal latency.

\paragraph*{History encoding}
At fixed $T=16$ and $K=32$, Table~\ref{tab:history_encoding} compares 4/16/64-token allocations, ordered oldest to newest. With the same 16 frames and 256 tokens per view, the 8:6:2 hierarchy achieves 98.6\% success versus 87.5\% for uniform encoding; fewer dense frames and longer histories at this budget perform worse. Relative to dense encoding of all 16 frames, it retains comparable success (98.6\% versus 98.0\%) with 75\% fewer history tokens, reduces peak memory from 38.3 to 23.4\,GB, and cuts proposal latency from 1.4 to 1.1\,s (21.4\%). Figure~\ref{fig:representation_diagnostics}(a) visualizes the DINOv3 spatial similarities supplied to compression.

\paragraph*{Joint generation}
Action only removes visual supervision and output. Visual auxiliary retains $K=32$ visual supervision of shared parameters, blocks action queries from visual states in every flow layer throughout training, and generates only actions at inference. Success rises from 93.6\% to 96.4\% with auxiliary supervision (+2.8 points), then to 98.6\% with joint generation (+2.2 points), supporting coupled training and sampling (Table~\ref{tab:joint_foresight}(a)). Auxiliary/joint proposal costs are 0.8/1.1\,s and 11.6/23.4\,GB. Forecasts are compared with observations from executing paired actions from the same simulator state, aligned by camera and physical time. PSNR measures framewise pixel error in dB; S3D-based FVD compares video-feature distributions affected by appearance and motion. SSIM/LPIPS are scaled by 100. PACT-WAM has similar PSNR to RynnVLA-002 (21.82/21.88\,dB), with higher SSIM and lower LPIPS/FVD (Table~\ref{tab:joint_foresight}(b)); Figs.~\ref{fig:sim_rollout} and~\ref{fig:real_rollout}(b) show rollouts.

\paragraph*{Visual capacity}
The $K=64$ variant improves all four forecast metrics over the default, but achieves 97.3\% control success versus 98.6\% at $K=32$ (Table~\ref{tab:joint_foresight}). Higher forecast fidelity thus does not necessarily improve manipulation success. Compared with $K=64$, the default halves visual positions, saves 0.3\,s per proposal, and reduces peak memory from 27.8 to 23.4\,GB. We use $K=32$ to balance control success and proposal cost. Figure~\ref{fig:representation_diagnostics}(b) illustrates the codec's reconstruction at this capacity.

\paragraph*{Proposal review}
Reviewer diagnostics use 5,000 default-$K=32$ proposals, one random valid decision from each of 100 w/o-PR episodes per task, including successes and failures; an independent 500-proposal pilot is excluded. All inputs share actions, saved states, and execution labels from 16-step replay. Prefix qualification follows predefined execution criteria without requiring task completion. Each condition has five technical-exception batches (0.10\%; 99.90\% coverage). The common valid set has 4,985 proposals (99.70\%), comprising 2,000 failed and 17,940 qualified prefixes. FA/FR evaluate the four reviews and controller jointly: across all $h\in\mathcal P$, FA divides failed prefixes with $k\ge h$ by all failed prefixes, and FR divides qualified prefixes with $k<h$ by all qualified prefixes. Unknown remains included through $k$; offline candidates are never resampled. Predicted previews reduce both rates (Table~\ref{tab:review_analysis}(a)).

Over 5,000 episodes per protocol, PR reaches 92.8\% success: 0.9 percentage points above full-chunk execution and 0.4 above Fixed-6 (Table~\ref{tab:review_analysis}(b)). All generate 16-step proposals. PR averages 14.0 approved steps over decisions with $k>0$; fixed horizons denote configured prefixes before replanning, independent of early task termination. These ablations support PACT-WAM's design: hierarchical encoding preserves temporal coverage at lower cost, joint generation improves control success, and compact visual latents balance success and proposal cost. Predicted previews improve review accuracy, while PR further improves closed-loop success.

% Keep the experimental tables with the experiment section.
\FloatBarrier
\section{Conclusion}
\label{sec:conclusion}

PACT-WAM combines hierarchical history encoding, compact visual latents, and coupled flow sampling to generate 16 paired actions and multi-view forecasts. It retains 16 observations with 75\% fewer history tokens than dense encoding. The base policy performs competitively on LIBERO, RoboTwin 2.0, and Piper, while decoded forecasts support additional test-time gains through PR. Ablations quantify the benefits and costs of history allocation, joint generation, visual capacity, and forecast-guided execution.

% \begingroup
% \clubpenalty=10000
% \noindent\textbf{Limitations and Future Work.}
\paragraph*{Limitations and Future Work}
PACT-WAM uses a fixed recency-based history allocation, which does not adapt visual detail to the task relevance of past observations. Its fixed 16-step prediction horizon limits the temporal scope of each proposal. PR relies on forecast accuracy and reviews only selected checkpoints, so brief failures between previews may escape detection. Future work will explore task-adaptive history compression, variable-horizon generation, and uncertainty-aware review with finer checks around critical transitions.
% \par\endgroup

% Start the bibliography after the main text; keep the final columns balanced.
\clearpage
\balance
\bibliographystyle{IEEEtran}
\bibliography{reference_style,references}

\end{document}